\documentclass[runningheads]{llncs}
\usepackage{graphicx}

\usepackage{tikz}
\usepackage{comment} 
\usepackage{amsmath,amssymb} 
\usepackage{color}

\usepackage{multirow}
\newcommand{\tabincell}[2]{\begin{tabular}{@{}#1@{}}#2\end{tabular}}
\usepackage{booktabs}
\usepackage{subfigure}
\usepackage{floatrow}
\newfloatcommand{capbtabbox}{table}[][\FBwidth]

\begin{document}
\pagestyle{headings}
\mainmatter
\def\ECCVSubNumber{2612}  

\title{Guessing State Tracking for Visual Dialogue} 

%
\author{Wei Pang \and
Xiaojie Wang}
\authorrunning{Wei Pang and Xiaojie Wang}
%
\institute{Center for Intelligence Science and Technology, School of Computer Science,\\
Beijing University of Posts and Telecommunications\\
\email{\{pangweitf,xjwang\}@bupt.edu.cn}}
\maketitle

\begin{abstract}
The Guesser is a task of visual grounding in GuessWhat?! like visual dialogue. It locates the target object in an image supposed by an Oracle oneself over a question-answer based dialogue between a Questioner and the Oracle. Most existing guessers make one and only one guess after receiving all question-answer pairs in a dialogue with the predefined number of rounds. This paper proposes a guessing state for the Guesser, and regards guess as a process with change of guessing state through a dialogue. A guessing state tracking based guess model is therefore proposed. The guessing state is defined as a distribution on objects in the image. With that in hand, two loss functions are defined as supervisions for model training. Early supervision brings supervision to Guesser at early rounds, and incremental supervision brings monotonicity to the guessing state. Experimental results on GuessWhat?! dataset show that our model significantly outperforms previous models, achieves new state-of-the-art, especially the success rate of guessing 83.3\% is approaching the human-level accuracy of 84.4\%.
\end{abstract}

\section{Introduction}

\begin{figure}[ht!]
\centering
\includegraphics[width=1.0\columnwidth]{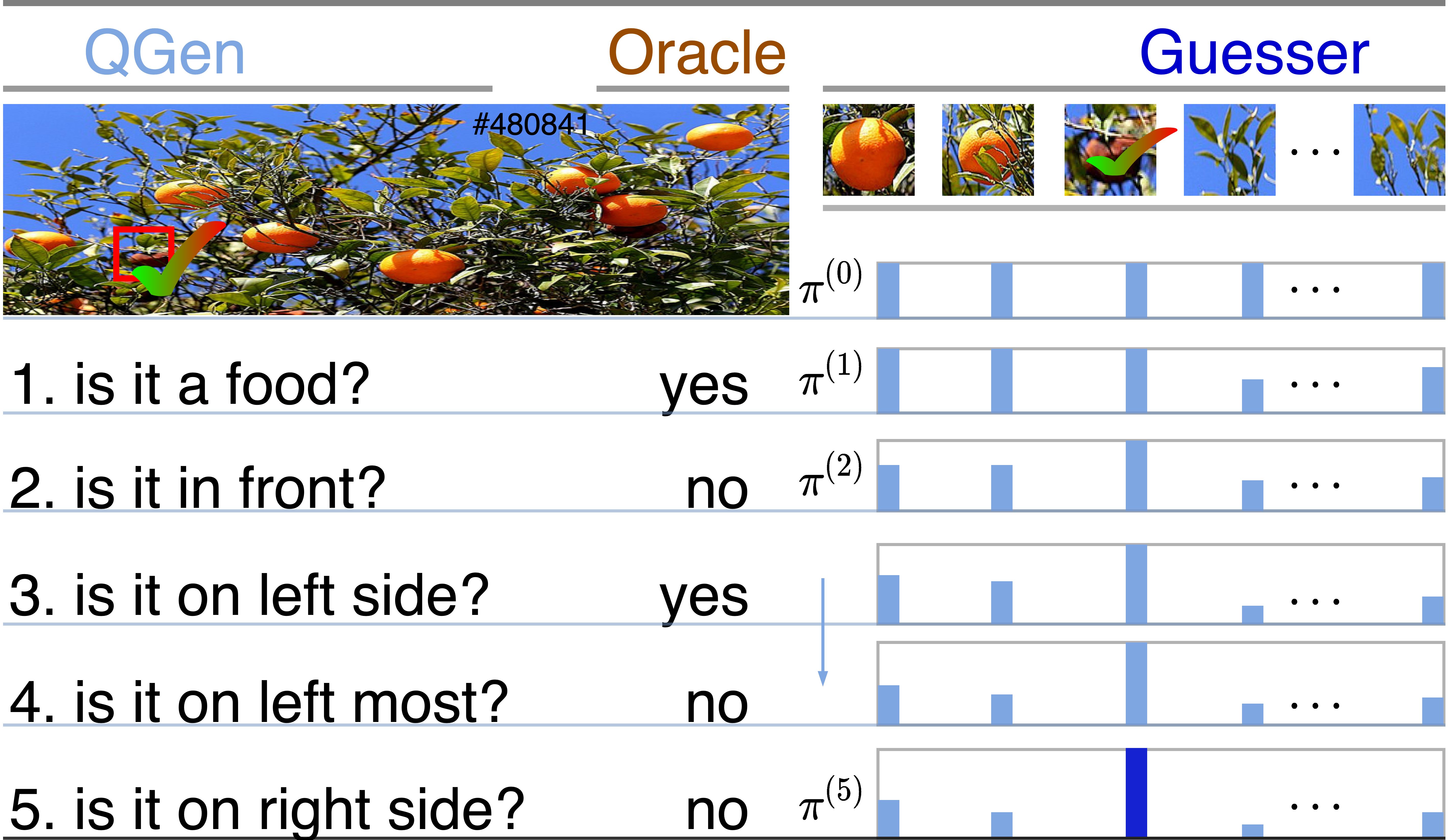}
\caption{The left part shows a game of GuessWhat?!. The right part illustrates the guess in Guesser as a process instead of a decision in a single point (the strips lineup denotes a probability distribution over objects, the arrowhead represents the tracking process).}\label{fig_illustration}
\end{figure}
Visual dialogue has received increasing attention in recent years. It involves both vision and language processing and interactions between them in a continuous conversation and brings some new challenging problems. Some different tasks of visual dialogue have been proposed, such as Visual Dialog \cite{CVPR17Visdial}, GuessWhat?! \cite{CVPR17GuessWhat}, GuessWhich \cite{HCOMP17GuessWhich}, and MNIST Dialog \cite{GuessNumber18,MNISTDialog17,NIPS18AQM}. Among them, GuessWhat?! is a typical object-guessing game played between a Questioner and an Oracle. Given an image including several objects, the goal of the Questioner is to locate the target object supposed by the Oracle oneself at the beginning of a game by asking a series of yes/no questions. The Questioner, therefore, has two sub-tasks: one is Question Generator (QGen) that asks questions to the Oracle, the other is Guesser that identifies the target object in the image based on the generated dialogue between the QGen and Oracle. The Oracle answers questions with yes or no. The left part of Fig.\ref{fig_illustration} shows a game played by the QGen, Oracle, and Guesser. The Guesser, which makes the final decision, is the focus of this paper.

Compared with QGen, relatively less work has been done on Guesser. It receives as input a sequence of question-answer (QA) pairs and a list of candidate objects in an image. The general architecture for Guesser introduced in \cite{CVPR17GuessWhat,IJCAI17Strub} that encodes the QA pairs into a dialogue representation and encodes each object into an embedding. Then, it compares the dialogue representation with any object embedding via a dot product and outputs a distribution of probabilities over objects, the object with higher probability is selected as the target. Most current work focuses on encoding and fusing multiple types of information, such as QA pairs, images, and visual objects. For example, some models \cite{CVPR17GuessWhat,IJCAI17Strub,AAAI20VDST,ACL19QGen,ECCV183rewards,COLING18No,NAACL19Jointly,ECCV18Add} convert the dialogue into a flat sequence of QA pair handled by a Long Short-Term Memory (LSTM)\cite{LSTM1997}, some models \cite{IJCAI18TPG,CVPR19Uncertainty,CVPR2018VG,ICCV2019HAST} introduce attention and memory mechanism to obtain a multi-modal representation of the dialogue.

Most of the existing Guesser models make a guess after fixed rounds of QA pairs, and this does not fully utilize the information from the sequence of QA pairs, we refer to that way as single-step guessing. Different games might need different rounds of QA pairs. Some work \cite{COLING18No,CVPR19Uncertainty} has therefore been done on choosing when to guess, i.e., make a guess after different rounds of question-answer for different games.

No matter the number of question-answer rounds is fixed or changed in different dialogues, existing Guesser models make one and only one guess after the final round of question-answer, i.e., Guesser is not activated until it reaches the final round of dialogue.

This paper models the Guesser in a different way. We think the Guesser to be active throughout the conversation of QGen and Oracle, rather than just only guessing at the end of the conversation. It keeps on updating a guess distribution after each question-answer pair from the beginning and does not make a final guess until the dialogue reaches a predefined round or it can make a confident guess. For example, as shown in Figure \ref{fig_illustration}, a guess distribution is initiated as uniform distribution, i.e., each object has the same probability as the target object at the beginning of the game. After receiving the first pair of QA, the guesser updates the guess distribution and continues to update the distribution in the following rounds of dialogue. It makes a final guess after predefined five rounds of dialogue.

We think that modeling the Guesser as a process instead of a decision in a single point provides more chances to not only make much more detailed use of dialogue history but also combine more information for making better guesses. One such information is monotonicity, i.e., a good enough guesser will never reduce the guessing probability on the target object by making proper use of each question-answer pair. A good guess either raises the probability of a target object in guess distribution when the pair contains new information about the target object or does not change the probability when the pair contains no new information.

This paper proposes a guessing state tracking (GST) based Guesser model for implementing the above idea. Guessing state (GS) is at first time introduced into the game. A GS is defined as a distribution on candidate objects. A GST mechanism, which includes three sub-modules, is proposed to update GS after each question-answer pair from the beginning. Update of Visual Representation (UoVR) module updates the representation of image objects according to the current guessing state, QAEncoder module encodes the QA pair, and Update of Guessing State (UoGS) module updates the guessing state by combining both information from the image and QA. GST brings a series of GS, i.e., let the Guesser make a series of guesses during the dialogue.

Two loss functions are designed on making better use of a series of GS, or the introduction of GS into visual dialogue makes the two new loss functions possible. One is called early supervision loss that tries to lead GS to the target object as early as possible, where ground-truth is used to guide the guesses after each round of QA, even the guess after the first round where a successful guess is impossible at that time. The other is called incremental supervision loss that tries to bring monotonicity mentioned above to the probability of target object in the series of GS.

Experimental results show that the proposed model achieves new state-of-the-art performances in all different settings on GuessWhat?!.
To summarize, our contributions are mainly three-fold:
\begin{itemize}
\item We introduce guessing state into visual dialogue for the first time and propose a Guessing State Tracking (GST) based Guesser model, a novel mechanism that models the process of guessing state updating over question-answer pairs. 
\item We introduce two guessing states based supervision losses, early supervision loss, and incremental supervision loss, which are effective in model training.
\item Our model performs significantly better than all previous models, and achieves new state-of-the-art in all different settings on GuessWhat?!. The guessing accuracy of 83.3\% approaches the human's level of 84.4\%.
\end{itemize}

\section{Related Work}

Visual Grounding is an essential language-to-vision problem of finding the most relevant object in an image by a natural language expression, which can be a phrase, a sentence, or a dialogue. It has attracted considerable attention in recent years\cite{arXiv16VG,CVPR2017speaker,CVPR2017Weakly,CVPR2018VG,arXiv18QGen,VG161,VG162}, and has been studied in the Guesser task in the GuessWhat?!\cite{CVPR2018VG}. This paper focuses on grounding a series of language descriptions (QA pair) in an image gradually by dialoguing.

Most previous work views Guesser as making a single-step guess based on a sequence of QA pairs. In \cite{CVPR17GuessWhat,IJCAI17Strub,ECCV183rewards,ACL19QGen,NAACL19Jointly,COLING18No,NIPS18AQM,AAAI20VDST,ICCV2019HAST,ECCV18Add}, all the multi-round QA pairs are considered as a flat sequence and encoded into a single representation using either an LSTM or an HRED\cite{hredm15} encoder, each object is represented as an embedding encoded from their object category embedding and 8-d spatial position embedding. A score is obtained by performing a dot product between the dialogue representation and each object embedding, then followed a softmax layer on the scores to output distribution of probabilities over objects, the object with higher probability is chosen as the most relevant object. As we can see, only one decision is made by the Guesser.

Most of the guesser models explored to encode the dialogue of multi-round QA pairs in an effective way. For example, in \cite{IJCAI18TPG,CVPR19Uncertainty}, they integrate Memory and Attention into the Guesser architecture used in \cite{IJCAI17Strub}. Where the memory is consist of some facts that are separately encoded from each QA pairs, the image feature vector is used as a key to attend the memory. In \cite{CVPR2018VG}, an accumulated attention (ATT) mechanism is proposed. It fuses three types of information, i.e., dialogue, image, and objects, by three attention models. Similarly, \cite{ICCV2019HAST} proposed a history-aware co-attention network (HACAN) to encode the QA pairs. 

As we can see, the models, as mentioned above, all make a single-step guess at the time that the dialogue ended, these might be counterintuitive. Different from them, we consider the guess as a continuous process, and explicitly track the guessing states after every dialogue round. Compared with prior works, we refer to the proposed GST as multi-step guessing.

Our GST based Guesser model is related to the VDST\cite{AAAI20VDST} based QGen model. \cite{AAAI20VDST} proposed a well-defined questioning state for the QGen and implemented a suitable tracking mechanism through the dialogue. The crucial difference in tracking state is that the QGen requires to track changes on the representations of objects because it needs more detailed information concerning the attended objects for asking more questions, while the Guesser does not need it.


\section{Model: Guessing State Tracking}
\begin{figure}[ht!] \centering
\includegraphics[width=1.0\columnwidth]{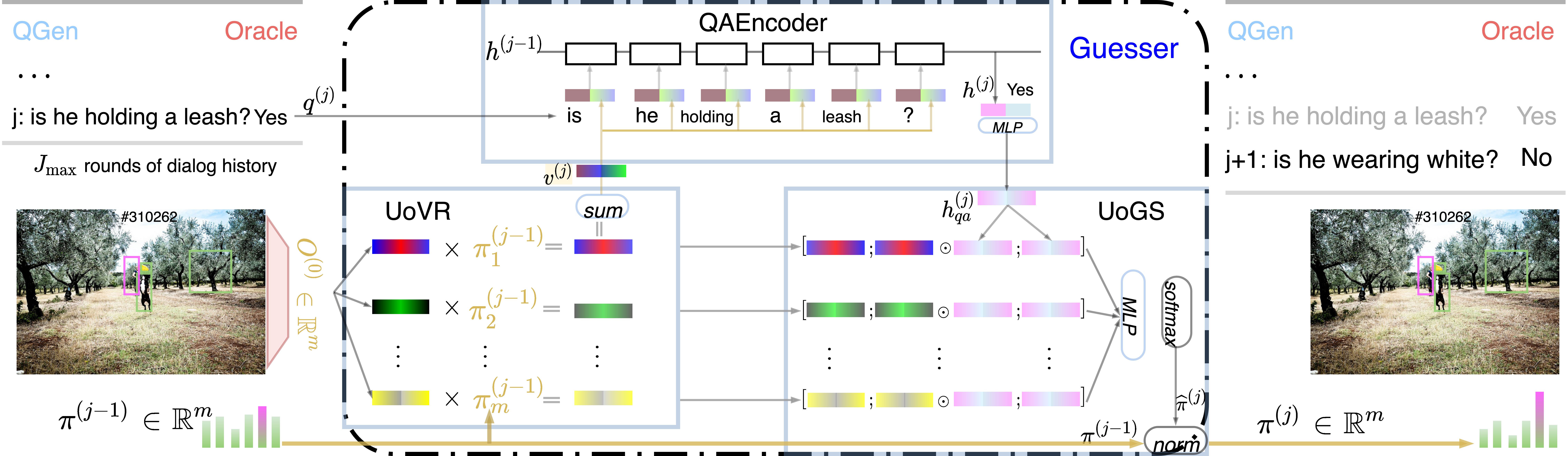}
\caption{Overview of the proposed Guesser model.}\label{fig_overview}
\end{figure}

The framework of our guessing state tracking (GST) model is illustrated in Fig.\ref{fig_overview}. Three modules are implemented in each round of guessing. There are Update of Visual Representation (UoVR), Question-Answer Encoder (QAEncoder), and Update of Guessing State (UoGS). Where, UoVR updates representation of an image for Guesser according to the previous round of guessing state, new visual representation is then combined into QAEncoder for synthesizing information from both visual and linguistic sides up to the current round of dialogue for the Guesser. Finally, UoGS is applied to update the guessing state of the guesser. We give details of each module in the following sub-sections.

\subsection{Update of Visual Representation (UoVR)}
Following previous work \cite{CVPR17GuessWhat,IJCAI17Strub}, candidate objects in an image are represented by their category and spatial features as in Eq.\ref{eqn_catespat}:
\begin{equation} \label{eqn_catespat}
\centering
O^{(0)} = \{o^{(0)}_{i} | o^{(0)}_{i} = \mathrm{MLP}([o_{cate}; o_{spat}])\}_{i=1}^{m},
\end{equation}
where $O^{(0)}\in \mathbb{R}^{m\times d}$ consists of m initial objects. For each object $o^{(0)}_{i}$, it is a concatenation of an 512-d category embedding $o_{cate}$ and an 8-d vector $o_{spat}$ of object location in an image. Where $o_{cate}$ are learnable parameters, $o_{spat}$ are coordinates $[x_{min},y_{min},x_{max},y_{max},x_{center},y_{center},w_{box},h_{box}]$ as in \cite{CVPR17GuessWhat}, $w_{box}$ and $h_{box}$ denote width and height of an object, the coordinates range from -1 to 1 scaled by the image width and height. To map object and word embedding to the same dimension, the concatenation is passed through an MLP to obtain a d-dimensional vector.

Let $\pi^{(j)}\in \mathbb{R}^{m}$ be an accumulative probability distribution over m objects after jth round of dialogue. It is defined as the guessing state and will be updated with the guessing process. At the beginning of a game, $\pi^{(0)}$ is a uniform distribution. With the progress of guessing, the visual representation in guesser's mind would update accordingly. Two steps are designed. The first step is an update of representations of objects. Pang and Wang\cite{AAAI20VDST} use an effective representation update in the VDST model. We borrow it for our GST model, as written in Eq.\ref{eqn_UoOR}:
\begin{equation} \label{eqn_UoOR}
\centering
O^{(j)} = (\pi^{(j-1)})^{T}O^{(0)},
\end{equation}
where $O^{(j)} \in \mathbb{R}^{m\times d}$ is a set of m updated objects at round j. Second, the summed embedding of all objects in $O^{(j)}$ is used as new visual representation as shown in Eq.\ref{eqn_sum},
\begin{equation} \label{eqn_sum}
\centering
v^{(j)} = \mathrm{sum}(O^{(j)}),
\end{equation}
where $v^{(j)} \in \mathbb{R}^{d}$ denotes updated visual information for the guesser at round j.

\subsection{Question-Answer Encoder (QAEncoder)}
For encoding linguistic information in the current question with visual information in hand, we concatenate $v^{(j)}$ to each word embedding $w^{(j)}_{i}$ in jth turn question $q^{(j)}$, take the concatenation as input to a single-layer LSTM encoder one by one as shown in Eq.\ref{eqn_LSTM},
\begin{equation} \label{eqn_LSTM}
\centering
h^{(j)} = \mathrm{LSTM}([w^{(j)}_{i}; v^{(j)}]_{i=1}^{N^{(j)}}, h^{(j-1)}),
\end{equation}
where $N^{(j)}$ is the length of question $q^{(j)}$. The last hidden state of the LSTM, $h^{(j)}$, is used as representation of $q^{(j)}$, and $h^{(j-1)}$ is used as initial input of the LSTM as shown in Fig.\ref{fig_overview}. 

$h^{(j)}$ is then concatenated to $a^{(j)}$, which is the embedding of the answer to jth turn question, the result $[h^{(j)}; a^{(j)}]$ are passed through an MLP to obtain the representation of QA pair at round j, as written in Eq.\ref{eqn_LMLP},
\begin{equation} \label{eqn_LMLP}
\centering
h^{(j)}_{qa} = \mathrm{MLP}([h^{(j)}; a^{(j)}]),
\end{equation}
where $h^{(j)}_{qa} \in \mathbb{R}^{d}$ synthesizes information from both questions and answers up to jth round dialogue for the guesser. It will be used to update the guessing state in the next module.

\subsection{Update of Guessing State (UoGS)}
When a new QA pair is received from the QGen and the Oracle, the Guesser needs to make a decision on which object would be ruled out, or which one would be gained more confidence, then renews its guessing state over objects in the image. Three steps are designed for updating the guessing state.

First, to fuse two types of information from QA and visual objects, we perform an element-wise product of $h^{(j)}_{qa}$ and each object embedding in $O^{(j)}$ to generate a joint feature for any object, as shown in Eq.\ref{eqn_attended},
\begin{equation} \label{eqn_attended}
\centering
O^{(j)}_{qa} = h^{(j)}_{qa} \odot O^{(j)},
\end{equation}
where $\odot$ denotes element-wise product, $O^{(j)}_{qa} \in \mathbb{R}^{m\times d}$ contains m joint feature objects.

Second, to measure how much the belief changes on ith object after jth dialog round, three feature vectors: the QA pair feature, joint feature of the ith object and updated representation of the ith object are concatenated together, and passed through a two-layers linear with a tanh activation, followed a softmax layer to produce a change of belief as described in Eq.\ref{eqn_CMM},
\begin{equation} \label{eqn_CMM}
\centering
\hat{\pi}^{(j)}_{i} = \mathrm{softmax}(W_{2}^{T}(tanh(W_{1}^{T}([h^{(j)}_{qa}; (O^{(j)}_{qa})_{i}; (O^{(j)})_{i}])))),
\end{equation}
where $i \in \{1,2,\dots,m\}$, $W_{1}\in \mathbb{R}^{1536\times 128}$ and $W_{2}\in \mathbb{R}^{128\times 1}$ are learnable parameters, the bias b is omitted in the linear layer for simplicity. $\hat{\pi}^{(j)}_{i} \in [0,1]$ means the belief changes on ith object after jth round. We find that this type of symmetric concatenation $[;;]$ in Eq.\ref{eqn_CMM}, where language and visual information are in a symmetrical position, is an effective way to handle multimodal information, which is also used in \cite{aaai20MBM}.

Finally, the previous rounds of guessing state $\pi^{(j-1)}$ are updated by multiplying $\hat{\pi}^{(j)}\in \mathbb{R}^{m}$, as follows: $\pi^{(j)} = norm(\pi^{(j-1)}\odot \hat{\pi}^{(j)})$. Where $\pi^{(j)}\in \mathbb{R}^{m}$ is the accumulated guessing state till round j, $norm$ is a sum-normalization method to make it a valid probability distribution, e.g., by dividing the sum of it.

\subsection{Early and Incremental Supervision}
 
The introduction of guessing states provides useful information for model training. Because the guessing states are tracked from the beginning of a dialogue, supervision of correct guess can be employed from an early stage, which is called early supervision. Because the guessing states are tracked at each round of dialogue, the change of guessing state can also be supervised to ensure that the guessing is alone in the right way. We call this kind of supervision, incremental supervision. Two supervision functions are introduced as follows.

{\bf Early Supervision}
Early supervision (ES) tries to maximize the probability of the right object from the beginning of a dialogue and keeps on using up to the penultimate round of the dialogue. It is defined as the summary of a series of cross-entropy between the guessing state and the ground-truth. That is:
\begin{equation} \label{eqn_esfloss}
\centering
L_{ES} = \frac{1}{J_{max}-1}\sum_{j=1}^{J_{max}-1}\mathrm{CrossEntropy }(\pi^{(j)}, y^{GT}),
\end{equation}
where $y^{GT}$ is a one-hot vector with 1 in the position of the ground-truth object, $J_{max}$ is the maximum number of rounds. The cross-entropy at the final round, i.e. $CrossEntropy(\pi^{(J_{max})}, y^{GT})$, we refer to as \textbf{plain supervision} loss ($L_{PS}$ in briefly).

{\bf Incremental Supervision}
Incremental supervision (IS) tries to keep the probability of the target object in guessing state increasing or nondecreasing as written in:
\begin{equation} \label{eqn_PRCFloss}
\centering
L_{IS} = -\sum_{j=1}^{J_{max}}\log(\pi^{(j)}_{target} - \pi^{(j-1)}_{target} + c),
\end{equation}
where $\pi^{(j)}_{target}$ denotes the target probability at round j. IS is defined as the change in probability to the ground-truth object before and after a round of dialog. Besides the log function that served as an extra layer of smooth, IS is somewhat similar to the progressive reward used in \cite{ECCV183rewards} that is from Guesser but as a reward for training QGen model. $c$ is a parameter that ensures the input to log be positive.

\subsection{Training}
Our model is trained in two stages, including supervised and reinforcement learning. For supervised learning, the guesser network is trained by minimizing the following objective as shown in Eq.\ref{eqn_loss},
\begin{equation} \label{eqn_loss}
\centering
L_{SL}(\theta) = \alpha (L_{ES} + L_{PS}) + (1 - \alpha) L_{IS},
\end{equation}
where $\alpha$ is a balancing parameter. For reinforcement learning, the guesser network is refined by maximizing the reward given in Eq.\ref{eqn_rlloss},
\begin{equation} \label{eqn_rlloss}
\centering
L_{RL}(\theta) = -E_{\pi_{\theta}}[\alpha (L_{ES} + L_{PS})+ (1 - \alpha) L_{IS})],
\end{equation}
where $\pi_{\theta}$ denotes a policy parameterized by $\theta$ which associates guessing state over actions, e.g., an action corresponds to select an object over m candidate objects. Following \cite{IJCAI18TPG}, we use the REINFORCE algorithm\cite{REINFORCE92} without baseline that updates policy parameters $\theta$.

\section{Experiments and Analysis}
\subsection{Experimental Setup}

{\bf Dataset} GuessWhat?! dataset containing 66k images, about 800k question-answer pairs in 150K games. It is split at random by 70\%, 15\%, 15\% of the games into the training, validation, and test set \cite{CVPR17GuessWhat,IJCAI17Strub}.

\noindent {\bf Baseline models} A GuessWhat?! game involves Oracle, QGen, and Guesser. Almost all existing work uses the same Oracle model \cite{CVPR17GuessWhat,IJCAI17Strub}, which will be used in all our experiments. Two different QGen models are used for validating our guesser model. One is the often used model in previous work \cite{IJCAI17Strub}, the other is a new QGen model which achieves new state-of-the-art \cite{AAAI20VDST}. Several different existing Guesser models are compared with our model. They are guesser \cite{CVPR17GuessWhat,IJCAI17Strub,arXiv18QGen}, guesser(MN) \cite{IJCAI18TPG,CVPR19Uncertainty}, GDSE \cite{NAACL19Jointly,COLING18No}, ATT \cite{CVPR2018VG} and HACAN \cite{ICCV2019HAST}. The models are first trained in a supervised way on the training set, and then, one Guesser and one QGen model are jointly refined by reinforcement learning or cooperative learning from self-play with the Oracle model fixed.

\noindent {\bf Implementation Details} The maximum round $J_{max}$ is set to 5 or 8 as in \cite{ECCV183rewards,ACL19QGen,AAAI20VDST}. The balancing parameter in Eq.\ref{eqn_loss} is set to 0.7, because we observe that our model achieves the minimum error rate on validation and test set when $\alpha = 0.7$. The parameter $c$ in Eq.\ref{eqn_PRCFloss} is set as 1.1. The size of word embedding and LSTM hidden unit number are set to 512. Early stopping is used on the validation set.

We use success rate of guessing for evaluation. Following previous work \cite{CVPR17GuessWhat,IJCAI17Strub,AAAI20VDST}, both success rates on NewObject and NewGame are reported. Results by three inference methods described in \cite{CVPR19Uncertainty}, including Sampling (S), Greedy (G) and Beam-search (BS, beam size is set to 20) are used on both NewObject and NewGame. Following \cite{IJCAI18TPG,CVPR19Uncertainty}, during joint reinforcement learning of Guesser and QGen models, only the generated successful games are used to tune the Guesser model, while all the generated games are used to optimize the QGen.

\noindent {\bf Supervised Learning (SL)} We separately train the Guesser and Oracle model for 20 epochs, the QGen for 50 epochs using Adam optimizer \cite{ICLR15Adam} with a learning rate of 3e-4 and a batch size of 64.

\noindent {\bf Reinforcement Learning (RL)} We use momentum stochastic gradient descent with a batch size of 64 with 500 epochs and learning rate annealing. The base learning rate is 1e-3 and decayed every 25 epochs with exponential rate 0.99. The momentum parameter is set to 0.9.

\begin{table}[htb!]
\begin{center}
\caption{Success rates of guessing (\%) with same Oracle (higher is better).} \label{tab_success}
\begin{tabular}{l|l|l|c|c|c|c|c|c|c}
\hline
&\multicolumn{2}{c|}{Questioner}&\multirow{2}{*}{Max Q's}&\multicolumn{3}{|c}{New Object} & \multicolumn{3}{|c}{New Game}\\ \cline{2-3} \cline{5-7} \cline{7-10}
&Guesser&QGen&&S & G & BS &S & G & BS\\
\hline
\multirow{7}{*}{\rotatebox{90}{pretrained in SL}}&guesser\cite{CVPR17GuessWhat}&qgen\cite{CVPR17GuessWhat}&5&41.6&43.5&47.1&39.2&40.8&44.6\\
&guesser(MN)\cite{IJCAI18TPG}&TPG\cite{IJCAI18TPG}&8&-&48.77&-&-&-&-\\
\cline{2-10}
&guesser\cite{IJCAI17Strub}&\multirow{2}{*}{qgen\cite{IJCAI17Strub}}&8& - & 44.6 & - &-&-&-\\
&GST(ours)&&8&41.73&\textbf{44.89}&-&39.97&41.36&-\\
\cline{2-10}
&\multirow{2}{*}{guesser\cite{IJCAI17Strub}}&\multirow{4}{*}{\tabincell{l}{VDST\cite{AAAI20VDST}}}&5&45.02&49.49&-&42.92&45.94&-\\
&&&8&46.70&48.01&-&44.24&45.03&-\\ \cline{2-2}
&\multirow{2}{*}{GST(ours)}&&5&\textbf{49.55}&\textbf{53.35}&\textbf{53.17}&\textbf{46.95}&\textbf{50.58}&\textbf{50.71}\\
&&&8&\textbf{52.71}&\textbf{54.10}&\textbf{54.32}&\textbf{50.19}&\textbf{50.97}&\textbf{50.99}\\
\hline
\multirow{2}{*}{\rotatebox{90}{SL}}&\multirow{2}{*}{GDSE-SL\cite{NAACL19Jointly}}&\multirow{2}{*}{GDSE-SL\cite{NAACL19Jointly}}&5&-&-&-&-&47.8&-\\
&&&8&-&-&-&-&49.7&-\\ \hline
\multirow{2}{*}{\rotatebox{90}{CL}}&\multirow{2}{*}{GDSE-CL\cite{NAACL19Jointly}}&\multirow{2}{*}{GDSE-CL\cite{NAACL19Jointly}}&5&-&-&-&-&53.7&-\\
&&&8&-&-&-&-&58.4&-\\ \hline
\multirow{2}{*}{\rotatebox{90}{AQM}}&\multirow{2}{*}{guesser\cite{NIPS18AQM}}&randQ\cite{NIPS18AQM}&5&-&-&-&-&42.48&-\\
&&countQ\cite{NIPS18AQM}&5&-&-&-&-&61.64&-\\ \hline
\multirow{15}{*}{\rotatebox{90}{trained by RL}}&\multirow{5}{*}{guesser(MN)\cite{IJCAI18TPG}}&\multirow{2}{*}{TPG\cite{IJCAI18TPG}}&5&62.6&-&-&-&-&-\\
&&&8&-&-&-&-&74.3&-\\ \cline{3-3}
&&ISM\cite{arXiv18QGen}&-&74.4&-&-&72.1&-&-\\
&&TPG\cite{IJCAI18TPG}&8& - & 74.3 & - &-&-&-\\
&&ISD\cite{CVPR19Uncertainty}&5& 68.3 & 69.2 & - &66.3&67.1&-\\ \cline{2-10}
&\multirow{5}{*}{guesser\cite{IJCAI17Strub}}&VQG\cite{ECCV183rewards}&5&63.2&63.6 & 63.9 &59.8&60.7&60.8\\
&&ISM\cite{arXiv18QGen}&-&-&64.2&-&-&62.1&-\\
&&ISD\cite{CVPR19Uncertainty}&5& 61.4 & 62.1 & 63.6 &59.0&59.8&60.6\\
&&RIG(rewards)\cite{ACL19QGen}&8&65.20&63.00&63.08&64.06&59.0&60.21\\
&&RIG(loss)\cite{ACL19QGen}&8&67.19&63.19&62.57&65.79&61.18&59.79\\
\cline{2-10}
&\multirow{2}{*}{guesser\cite{IJCAI17Strub}}&\multirow{5}{*}{\tabincell{l}{qgen\cite{IJCAI17Strub}}}&5& 58.5 & 60.3 & 60.2 &56.5&58.4&58.4\\
&&&8& 62.8 & 58.2&53.9 &60.8&56.3&52.0\\ \cline{2-2}
&\multirow{2}{*}{guesser(MN)\cite{IJCAI18TPG}}&&5&59.41&60.78&60.28&56.49&58.84&58.10\\
&&&8&62.05&62.73&-&59.04&59.50&-\\ \cline{2-2}
&GST(ours)&&5&\textbf{64.78}&\textbf{67.06}&\textbf{67.01}&\textbf{61.77}&\textbf{64.13}&\textbf{64.26}\\
\cline{2-10}
&\multirow{2}{*}{guesser\cite{IJCAI17Strub}}&\multirow{4}{*}{\tabincell{l}{VDST\cite{AAAI20VDST}}}&5&66.22&67.07&67.81&63.85&64.36&64.44\\
&&&8&69.51&70.55&71.03&66.76&67.73&67.52\\ \cline{2-2}
&\multirow{2}{*}{GST(ours)}&&5&\textbf{77.38}&\textbf{77.30}&\textbf{77.23}&\textbf{75.11}&\textbf{75.20}&\textbf{75.13}\\
&&&8&\textbf{83.22}&\textbf{83.32}&\textbf{83.46}&\textbf{81.50}&\textbf{81.55}&\textbf{81.62}\\
\hline
&Human\cite{IJCAI17Strub}&-&-&-&84.4&-&-&84.4&-\\
\hline
\end{tabular}
\end{center}
\end{table}

\subsection{Comparison with the state-of-the-art}

\noindent {\bf Task Success Rate} Table \ref{tab_success} reports the success rate of guessing with different combinations of QGen and Guesser models with the same Oracle model used in \cite{CVPR17GuessWhat,IJCAI17Strub} for the GuessWhat?! game.

In the first part of table \ref{tab_success}, all models are trained in SL way. We can see that no matter which QGen models are used, qgen \cite{IJCAI17Strub} or VDST \cite{AAAI20VDST}, our guesser model GST significantly outperforms other guesser models in both 5 and 8 rounds dialogue at all different settings. Specifically, GST achieves a new state-of-the-art of 54.10\% and 50.97\% on NewObject and NewGame in Greedy way by SL. 

In the second part of table \ref{tab_success}, two combinations trained in cooperative learning (CL) way are given. Our model is not trained in this way. So we do not have a comparison in CL case with the performance of these models are lower than those in the RL part. 

\begin{figure}[ht!]
\centering
\subfigure[]{
\label{fig_alpha:a}
\includegraphics[width=0.48\columnwidth]{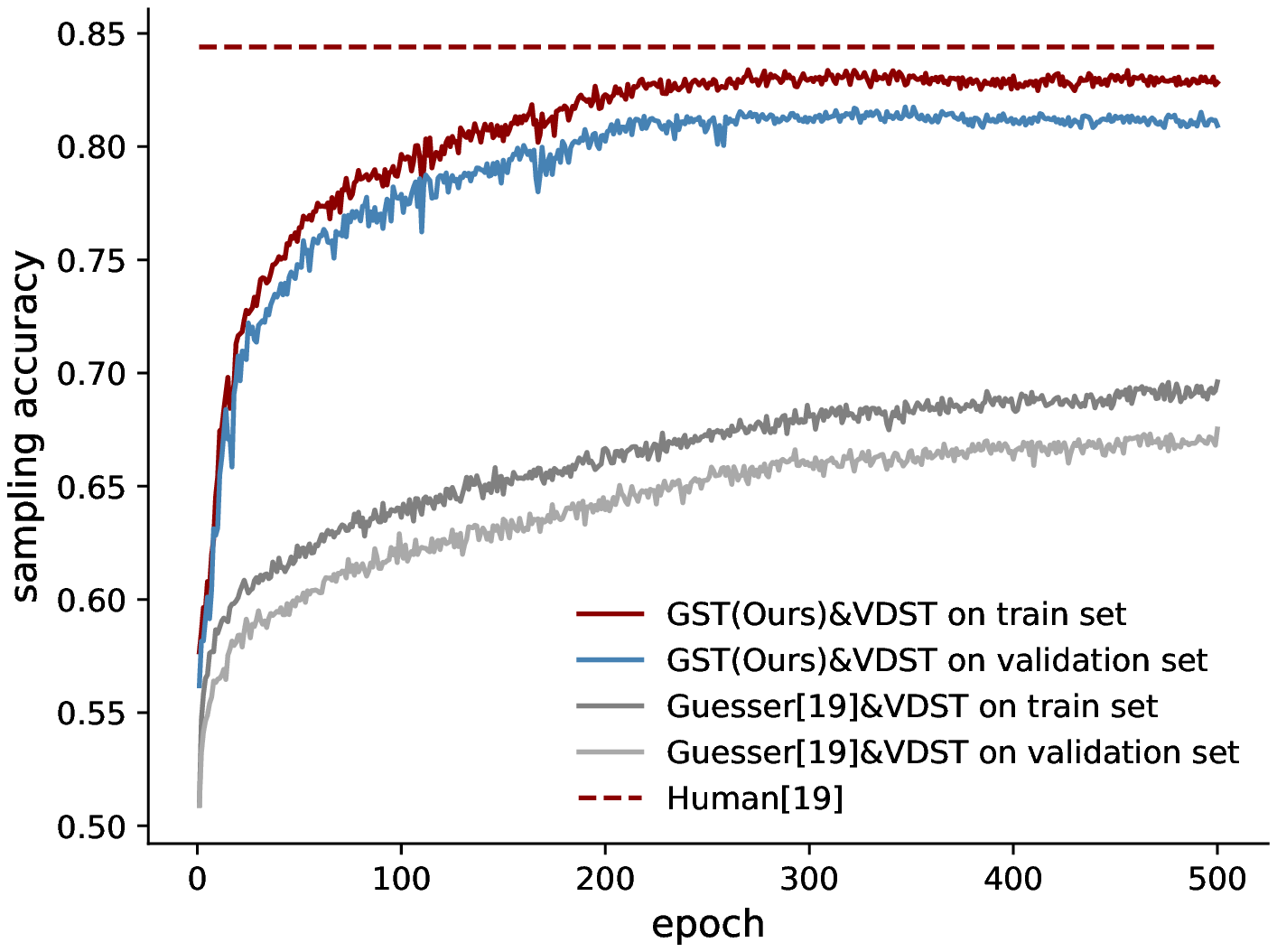}}
\subfigure[]{
\label{fig_alpha:b}
\includegraphics[width=0.48\columnwidth]{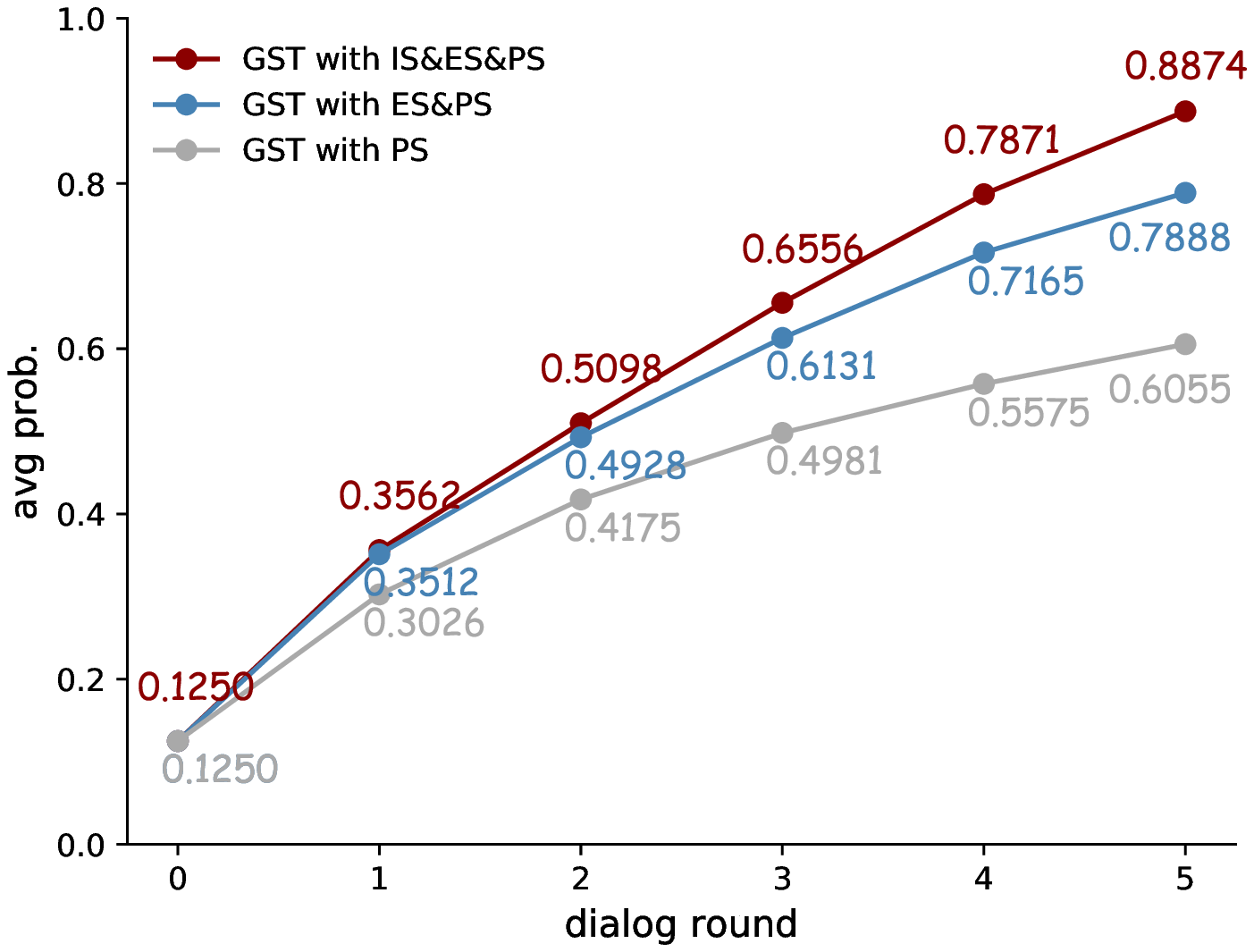}}
\caption{\textbf{a}, Sampling accuracy of reinforcement learning on training and validation set, our GST outperforms guesser\cite{IJCAI17Strub} by a large margin. \textbf{b}, Average belief of the ground-truth object at each round, changes with an increase in the number of dialogue rounds.}
\label{fig_alpha}
\end{figure}

In the third part of table \ref{tab_success}, all QGen and Guesser models are trained by RL. We can see that our GST Guesser model combined with the VSDT QGen model achieves the best performance in both 5 and 8 rounds dialogue at all different settings. It significantly outperforms other models. For example, it outperforms the best previous model at Sampling (S) setting on NewObject (i.e. guesser(MN)\cite{IJCAI18TPG} + ISM \cite{arXiv18QGen} with 72.1\%) by nearly 9 percent, outperforms the best previous model at Greedy (G) setting on NewObject (i.e. guesser(MN) \cite{IJCAI18TPG} + TPG \cite{IJCAI18TPG} with 74.3\%) by more than 9 percent, outperforms the best previous model in NewObject at Beam-search (BS) setting on NewObject (i.e. guesser \cite{IJCAI17Strub} + VDST \cite{AAAI20VDST} with 71.03\%) by more than 12 percent. The same thing happens on NewGame case. That is to say, our model consistently outperforms previous models in all different settings on both NewObject and NewGame. Especially, GST achieves 83.32\% success rate on NewObject in Greedy way, which is approaching human performance 84.4\%. Fig.\ref{fig_alpha:a} shows the learning curve for joint training of GST Guesser and VDST QGen with 500 epochs in Sampling way, it shows superior accuracy compared to the Guesser model\cite{IJCAI17Strub} trained with VDST QGen.

Specifically, with same QGen (no matter which QGen models are used, qgen used in \cite{IJCAI17Strub} or VDST used in \cite{AAAI20VDST}), our guesser model GST significantly outperforms other guesser models in both 5 and 8 rounds dialogue at all different settings. It demonstrates that GST is more able to ground a multi-round QA pairs dialogue in the image compared to previous single-step guessing models for the GuessWhat?! game.

\begin{table}[htb!]
\begin{center}
\caption{Error rate (\%) on the GuessWhat?! dataset (lower is better).} \label{tab_error}
\begin{tabular}{lcccc}
\hline
Model & Train err& Val err& Test err&\tabincell{l}{Max Q's}\\ \hline
Random\cite{CVPR17GuessWhat}& 82.9&82.9&82.9&- \\
LSTM\cite{CVPR17GuessWhat} &27.9&37.9&38.7&- \\
HRED\cite{CVPR17GuessWhat}&32.6&38.2&39.0&- \\
Guesser\cite{IJCAI17Strub}&-&-&36.2&- \\
LSTM+VGG\cite{CVPR17GuessWhat}&26.1&38.5&39.5&- \\
HRED+VGG\cite{CVPR17GuessWhat}&27.4&38.4&39.6&- \\
ATT-r2\cite{CVPR2018VG}&29.3&35.7&36.5&- \\
ATT-r3\cite{CVPR2018VG}&30.5&35.1&35.8&- \\
ATT-r4\cite{CVPR2018VG}&29.8&35.3&36.3&- \\
ATT-r3(w2v)\cite{CVPR2018VG}&26.7&33.7&34.2&- \\
Guesser\cite{ACL19QGen}&-&-&35.8&- \\
HACAN\cite{ICCV2019HAST}&26.1&32.3&33.2&- \\ \hline
\tabincell{l}{GST(ours,\\ trained in SL)}&24.7&33.7&34.3&- \\ \cline{1-1}
\tabincell{l}{GST(ours,\\ trained in RL)}&\textbf{22.7}&\textbf{23.1}&\textbf{24.7}&5 \\ \cline{1-1}
\tabincell{l}{GST(ours,\\ trained in RL)}&\textbf{16.7}&\textbf{16.9}&\textbf{18.4}&8 \\ \hline
Human\cite{CVPR17GuessWhat} & 9.0&9.2 & 9.2\\ \hline
\end{tabular}
\end{center}
\end{table}

\begin{table}[ht!]
\begin{center}
\caption{Comparison of success rate with different supervisions in SL.} \label{tab_ablationESIS}
\begin{tabular}{llcc}
\hline
\multirow{2}{*}{\#} &\multirow{2}{*}{Model} & \multicolumn{2}{c}{New Object} \\  \cmidrule(r){3-4}
&& S & G \\ \hline
1&GST with ES\&PS and IS (full)&52.71& 54.10\\
2&$\quad-$ES\&PS&37.10&42.58\\
3&$\quad-$IS&48.96&53.49\\
\cline{3-4}
\cline{3-4}
& &  \multicolumn{2}{c}{New Game} \\ \cline{3-4}
1&GST with ES\&PS and IS (full)& 50.19& 50.97\\
2&$\quad-$ES\&PS&34.41&39.48\\
3&$\quad-$IS&46.10&50.33\\
\hline
\end{tabular}
\end{center}
\end{table}

\noindent {\bf Error Rate} For a fair comparison of Guesser models alone, we follow the previous work\cite{CVPR2018VG,ICCV2019HAST} by measuring error rate on training, validation, and test set. In Table \ref{tab_error}, we can see that GST trained in SL is comparable to more complex attention algorithms, such as ATT\cite{CVPR2018VG} and HACAN\cite{ICCV2019HAST}. After reinforcement learning, GST model achieves a lower error rate than the compared models in both 5 and 8 rounds, especially at 8 rounds, it obtains error rate of 16.7\%, 16.9\%, and 18.4\%, respectively.

\subsection{Ablation Study}

\noindent {\bf Effect of Individual Supervision} In this section, we conduct ablation studies to separate contribution of supervisions: Plain Supervision (PS), Early Supervision (ES) and Incremental Supervision (IS).

Table \ref{tab_ablationESIS} reports the success rate of guessing after supervised learning. Removing ES\&PS from the full model, the game success rate significantly drops 11.52 (from 54.10\% to 42.58\%) and 11.49 (from 50.97\% to 39.48\%) points on NewObject and NewGame on Greedy case. Removing IS, the success rate drops 0.61 (from 54.10\% to 53.49\%) and 0.64 (from 50.97\% to 50.33\%), respectively. It shows that early supervision pair with ES\&PS contributes more than incremental supervision.

We then analyze the impact of supervision losses to guessing state. We train three GST models with RL using three different loss functions, i.e. PS, PS\&ES, and PS\&ES\&IS respectively and then count the averaged probability of the ground-truth object based on all the successful games in test set at each round. Fig.\ref{fig_alpha:b} shows three curves of averaged belief changing with rounds of dialogue. As is observed, we have three notes.

First, guess probability is progressively increasing in all three different losses. It demonstrates our core idea: thinking of the guess as a process instead of a single decision is an effective practical way to ground a multi-round dialogue in an image. Because GST based Guesser makes use of more detailed information in the series of guessing states (GS), i.e. the two losses.

Second, average probability in the blue line, trained with ES\&PS, is higher than that in the gray line (trained in PS alone), it demonstrates the effectiveness of early supervision loss.

Third, average probability in the red line, trained with IS\&ES\&PS, is better than that in the blue line, it shows incremental supervision gives further improvement to guess.

Overall, these results demonstrate the effectiveness of early supervision and incremental supervision. It is the combination of these supervisions that train GST based guesser model efficiently.

\begin{table}
\begin{center}
\begin{floatrow}
\capbtabbox{
\begin{tabular}{lccc}
\hline
Concat & Train err& Val err& Test err\\ \hline
$[h_{qa}^{(j)};h_{qa}^{(j)}\odot O^{(j)};O^{(j)}]$&\textbf{24.8}&\textbf{33.7}&\textbf{34.4} \\
$[h_{qa}^{(j)}\odot O^{(j)}]$&26.3&35.7&36.7 \\
$[h_{qa}^{(j)};O^{(j)}]$&27.3&36.5&37.8 \\ \hline
\end{tabular}
}{
\caption{Comparison of error rate (\%) for three types of concatenation during SL.}
\label{tab_ablationSymCate}
}
\capbtabbox{
\begin{tabular}{lccc}
\hline
c & Train& Val& Test\\ \hline
c=1.1&24.7&\textbf{33.7}&\textbf{34.3} \\
c=1.5&26.5&34.1&34.8 \\
c=2.0&\textbf{23.3}&34.0&34.8 \\ \hline
\end{tabular}
}{
\caption{Error rate of different c in Eq.\ref{eqn_PRCFloss} during SL.}
\label{tab_ablationCloss}
}
\end{floatrow}
\end{center}
\end{table}

\noindent {\bf Effect of Symmetric Concatenation} In table \ref{tab_ablationSymCate}, compared with symmetric concatenation appears in Eq.\ref{eqn_CMM}, average error rate increases 2.9 points on all three sets if $[h_{qa}^{(j)};O^{(j)}]$ used and increases 1.9 points if $[h_{qa}^{(j)}\odot O^{(j)}]$ used. It indicates that symmetric concatenation serves as a valuable part in Eq.\ref{eqn_CMM}.
\begin{figure}[htb!] \centering
\includegraphics[width=1.0\columnwidth]{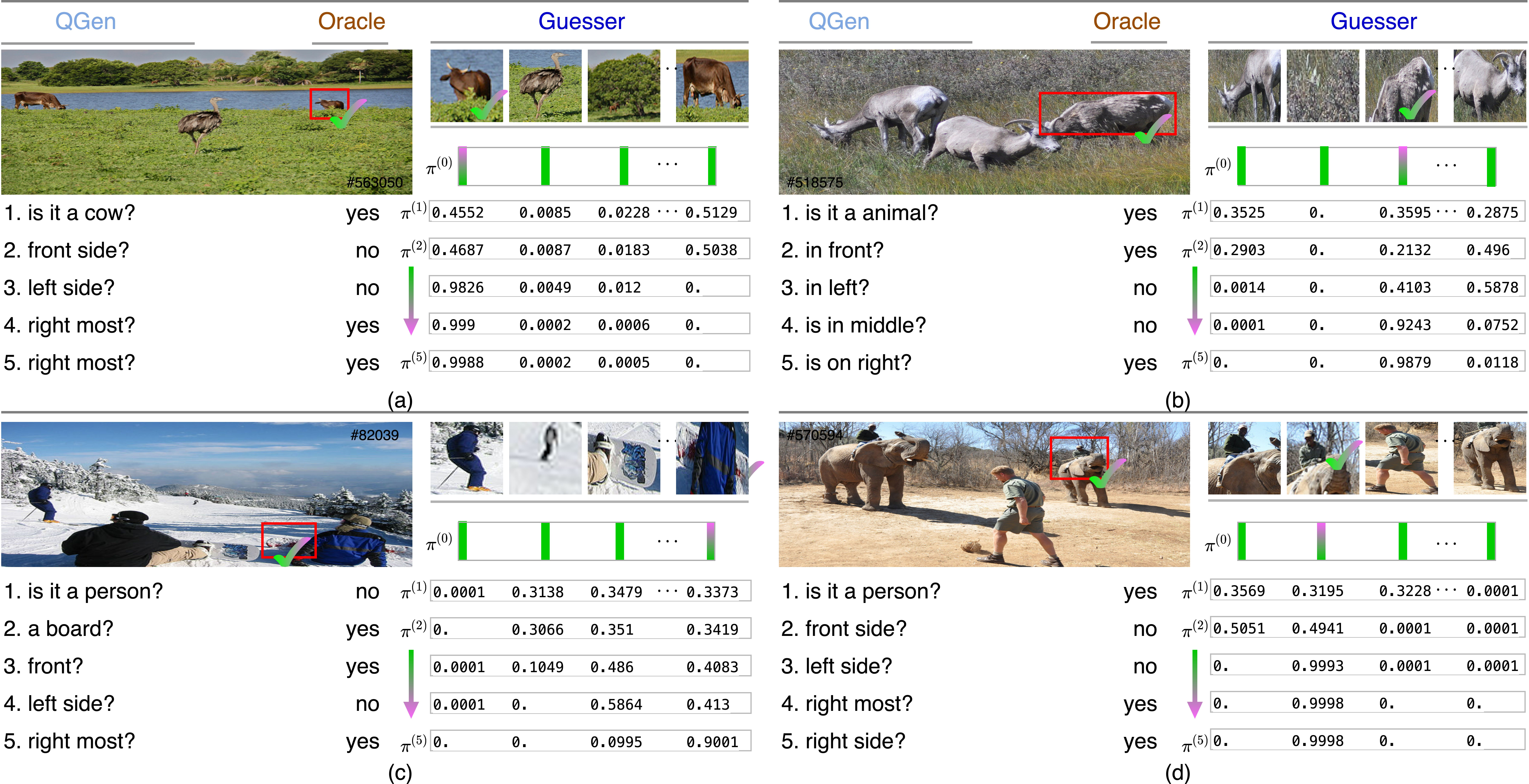}
\caption{Four successful games show the process of tracking guessing state.}\label{fig_examples}
\end{figure}

\noindent {\bf Effect of c in Eq.\ref{eqn_PRCFloss}}
Table \ref{tab_ablationCloss} shows error rate of different c in Eq.\ref{eqn_PRCFloss} trained with SL on three dataset. As is observed, c is insensitive to the error rate. We set c to 1.1 as it obtains a lower error rate on Val err and Test err.

\subsection{Qualitative Evaluation}
In Fig.\ref{fig_examples}, we show four successful dialogues to visualize the process in guessing. We plot 4 candidate objects for simplicity, $\pi^{(0)}$ represents a uniform distribution of initial guessing state, $\pi^{(1)}$ to $\pi^{(5)}$ show the process of tracking GS. Taking Fig.\ref{fig_examples}(a) as an example. Guesser has an initial uniform guess on all candidates, i.e. $\pi^{(0)}$. QGen starts a dialogue by asking “is it a cow?”, Oracle answer “yes”, then Guesser renews its $\pi^{(0)}$ to $\pi^{(1)}$. Specifically, the probabilities on the ostrich and tree approaches go down to close to 0, the cow on both sides increases to 0.45 and 0.51 respectively. At last, all the probabilities are concentrated on the cow on the right with high confidence of 0.9988, which is the guessed object. In \ref{fig_examples}(b) to \ref{fig_examples}(d), three more success cases are shown.

\subsection{Discussion on Stop Questioning}
When to stop questioning is also a problem in GuessWhat?! like visual dialogue. Most of the previous work chooses a simple policy, i.e., a QGen model stops questioning after a predefined number of dialogue rounds, and the guessing model selects an object as the guess. 

Our model can implement this policy by making use of $\pi^{(j)}$, the guessing state after the jth round dialogue. If $K$ is the predefined number, the guesser model will keep on updating $\pi^{(j)}$ till $j=K$. The object with the highest probability in $\pi^{(K)}$ will be then selected as the guess.
 
A same number of questions are asked for any game under this policy, no matter how different the different games are. The problem of the policy is obvious. On the one hand, the guesser model does not select any object even if it is confident enough about a guess and make a QGen model keep on asking till K questions are asked. On the other hand, the QGen model cannot ask more questions when K questions are asked even if the guesser model is not confident about any guess at that time. The guesser model must give a guess.

Our model can provide a chance to adopt some other policies for stopping questioning. A simple way is to predefine a threshold of confidence. Once the biggest probability in a guessing state is equal to or bigger than the threshold, question answering is stopped, and the guesser model output the object with the biggest probability as the guess. Another way involves the gain of guessing state. Once the information gain from the jth state to the j+1th state is less than a threshold, the guesser model outputs the object with the biggest probability as the guess.

\section{Conclusion}
The paper proposes a novel guessing state tracking (GST) based model for the Guesser, which models guess as a process with change of guessing state, instead of making one and only one guess, i.e. a single decision, over the dialogue history in all the previous work. To make full use of the guessing state, two losses, i.e., early supervision loss and incremental supervision loss, are introduced. Experiments show that our GST based guesser significantly outperforms all of the existing methods, and achieves new strong state-of-the-art accuracy that closes the gap to humans, the success rate of guessing 83.3\% is approaching the human-level accuracy of 84.4\%.

\clearpage

\section*{Acknowledgements}
We thank the reviewers for their comments and suggestions. This paper is supported by NSFC (No. 61906018), Huawei Noah's Ark Lab and MoE-CMCC “Artificial Intelligence” Project (No. MCM20190701).

%
%
\bibliographystyle{splncs04}
\bibliography{egbib}

\end{document}